\documentclass[fleqn,10pt]{wlscirep}
\usepackage[utf8]{inputenc}
\usepackage[T1]{fontenc}
\title{Markov versus quantum dynamic models of belief change during evidence
monitoring}

\author[1,*]{Jerome R. Busemeyer}
\author[1]{Peter D. Kvam}
\author[1,2,+]{Timothy J. Pleskac}
\affil[1]{Indiana University, Psychological Brain Sciences, Bloominton, 47405, USA}
\affil[2]{University of Kansas, Psychology, Lawrence, 66044, USA}

\affil[*]{jbusemey@indiana.edu}

\begin{abstract}
Two different dynamic models for belief change during evidence monitoring
were evaluated: Markov and quantum. They were empirically tested with
an experiment in which participants monitored evidence for an initial
period of time, made a probability rating, then monitored more evidence,
before making a second rating. The models were qualitatively tested
by manipulating the time intervals in a manner that provided a test
for interference effects of the first rating on the second. The Markov
model predicted no interference whereas the quantum model predicted
interference. A quantitative comparison of the two models was also
carried out using a generalization criterion method: the parameters
were fit to data from one set of time intervals, and then these same
parameters were used to predict data from another set of time intervals.
The results indicated that some features of both Markov and quantum
models are needed to accurately account for the results.
\end{abstract}
\begin{document}

\flushbottom
\maketitle
%
%
\thispagestyle{empty}


\section*{Introduction}
When monitoring evidence during decision making, a person's belief
about each hypothesis changes and evolves across time. For example,
when watching a murder mystery film, the viewer's beliefs about guilt
or innocence of each suspect change as the person monitors the evidence
from the ongoing movie. What are the basic dynamics that underlie
these changes in belief during evidence accumulation? Here we investigate
two fundamentally different ways to understand the dynamics of belief
change. 

\begin{figure}
\caption{\label{fig:Models} Illustration of Markov (Left) and quantum (Right)
processes for evolution of beliefs during evidence monitoring. The
horizontal axis represents 101 belief states associated with subjective
evidence scale values ranging from 0 to 100 in one unit steps. The
vertical axis represents time that has passed during evidence monitoring.
The Markov process moves from a belief state located at one value
at one moment in time to another belief state at a later moment to
produce a sample path across time. At the time of a request for a
rating, the current belief state is mapped to a probability rating
value (small circle). The quantum process assigns a distribution across
the belief states at one moment, which moves to another distribution
at a later moment (technically, this figure shows the squared magnitude
of the amplitude at each state). At the time of a request for a rating,
the current distribution is used to probabilistically select a rating
(red vertical line).}
\includegraphics[width=.5\textwidth]{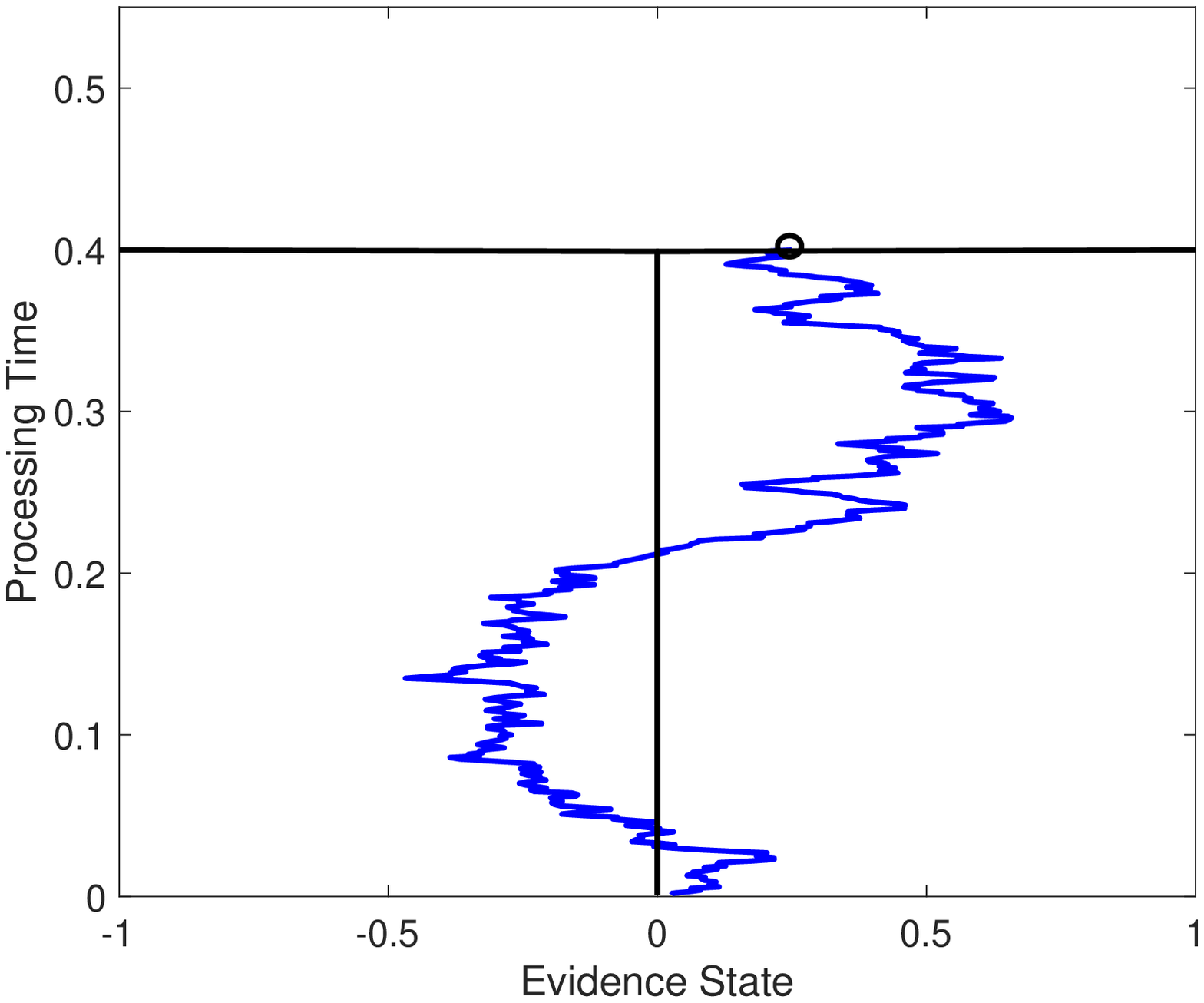}\includegraphics[width=.5\textwidth]{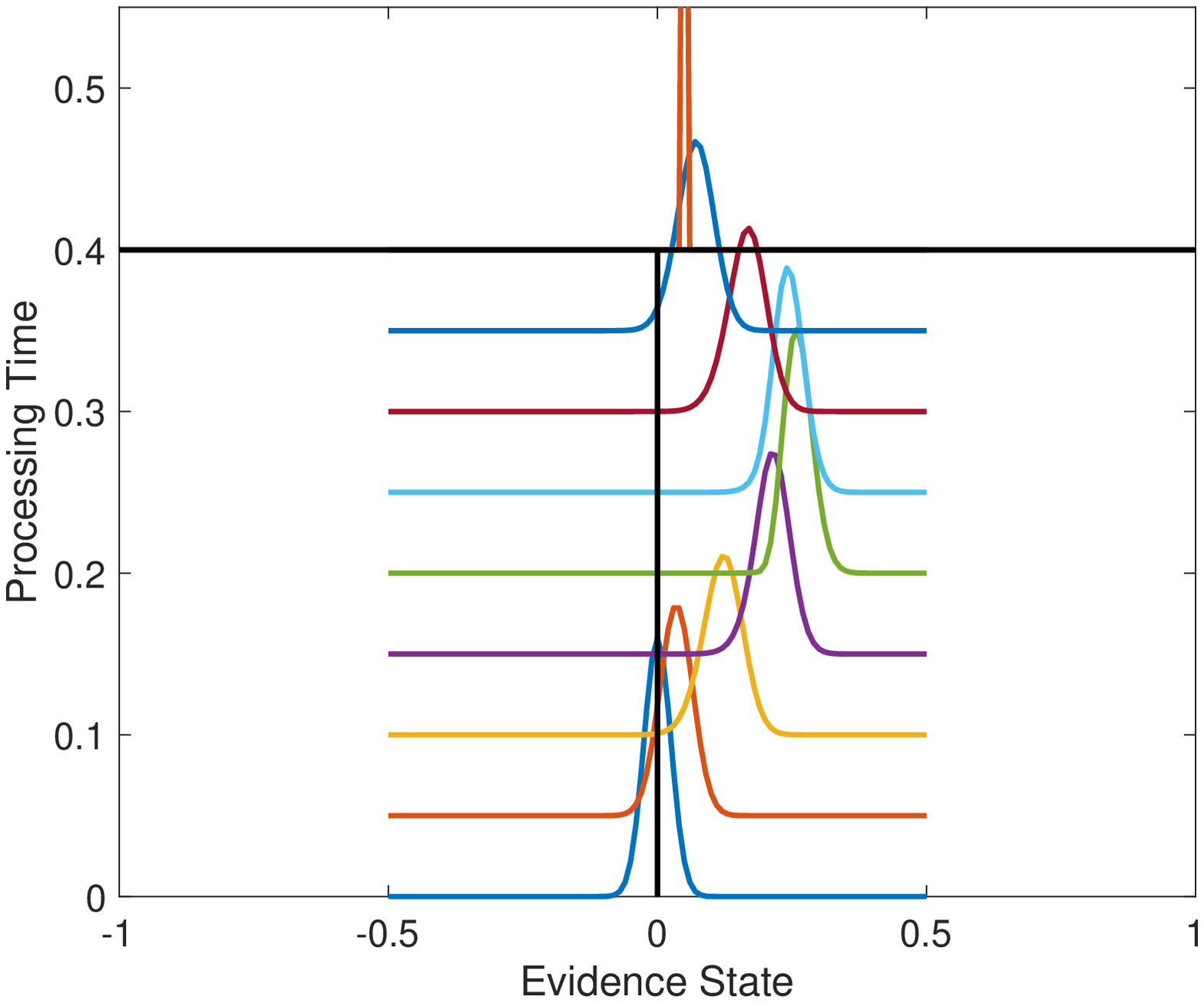}
\end{figure}

The ``classical\textquotedbl{} way of modeling evidence dynamics is
to assume that the dynamics follow a Markov process, such as a random
walk or a drift diffusion model (see, e.g., \cite{link1975relative, RatcliffTiCS2016, ShadlenKiani2013}). Note that
these models are essentially neuro-cognitive versions of a Bayesian
sequential sampling model in which the belief state at each moment
corresponds to the posterior probability of the accumulated evidence
\cite{bogacz2006physics}. According to one application of this view
\cite{PleskacBus2010,yu2015dynamics,kvam2016strength}, the decision
maker's belief state at any single moment is located at a specific
point on some mental scale of evidence. This belief state changes
moment by moment from one location to another on the evidence scale,
sketching out a sample path as illustrated in Figure \ref{fig:Models},
left panel. At the time point that an experimenter requests a probability
rating, the decision maker simply maps the pre-existing mental belief
state onto an observed rating response.

A ``non-classical'' way of modeling evidence dynamics (or belief change)
is to assume that the dynamics follow a quantum process \cite{Busemeyer:Wang:Townsend:2006,BusemeyerBruza2012,martinez2016quantum}.
According to one application of this view \cite{KvamPNAS2015}, the
decision maker's belief state at a moment is not located at any specific
point on the mental evidence scale; instead, at any moment, the belief
state is indefinite, which is represented by a superposition of beliefs
over the mental evidence scale. This superposition state forms a wave
that flows across time as illustrated in Figure \ref{fig:Models},
right panel (technically, this figure shows the squared magnitude
of the amplitude at each state). At the time point that an experimenter
requests a judgment, the judge's indefinite state must be resolved,
and the wave is ``collapsed'' to probabilistically select an observed
rating response (red bar at the top of Figure 1, right panel). Note
that we are not proposing that the brain is some type of quantum computer
-- we are simply using the basic principles of quantum probability
theory to predict human behavior. One possible neural network model
that could implement these principles was proposed by \cite{busemeyer2017neural}.

\begin{figure}
\caption{\label{fig:Task1}Illustrations of dot motion task with probability
rating scale and protocol for experiment (Left), timing of judgments
(right). On the left, a judge watches the dot motion and at the appointed
time selects a rating response from the semicircle. In the middle,
the judge first watches the dots motion for a period of time, then
makes a first choice response, and then observes the dot motion for
a period of time again, and then makes a second rating response. On
the right, each condition had a different pair of time points for
requests for ratings: the time intervals of the first two conditions
are contained within condition 3. Conditions 1 and 2 were used to
estimate model parameters and then these same model parameters were
used to predict the ratings for condition 3.}

\raggedright{}\includegraphics[width=.5\textwidth]{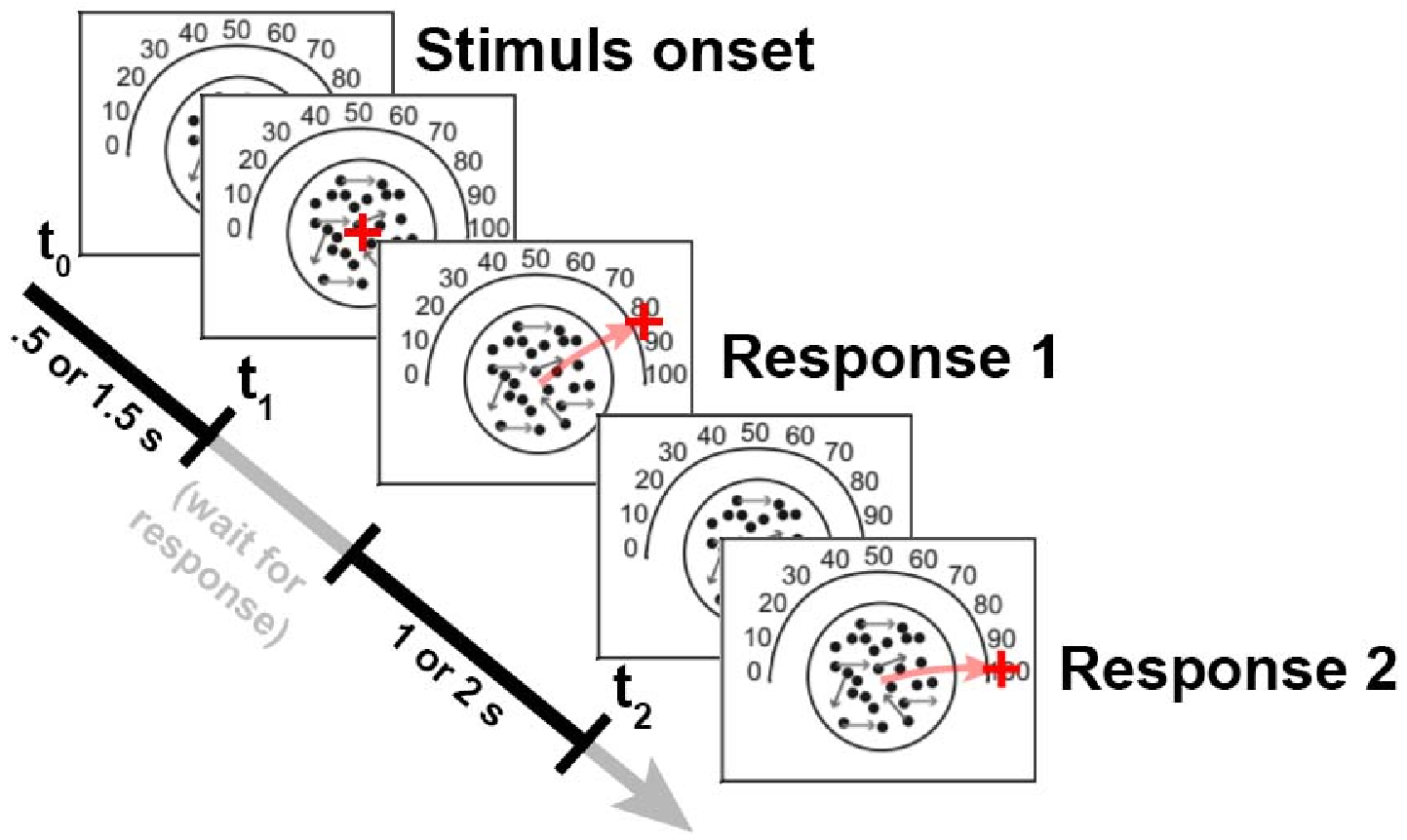}\includegraphics[width=.5\textwidth]{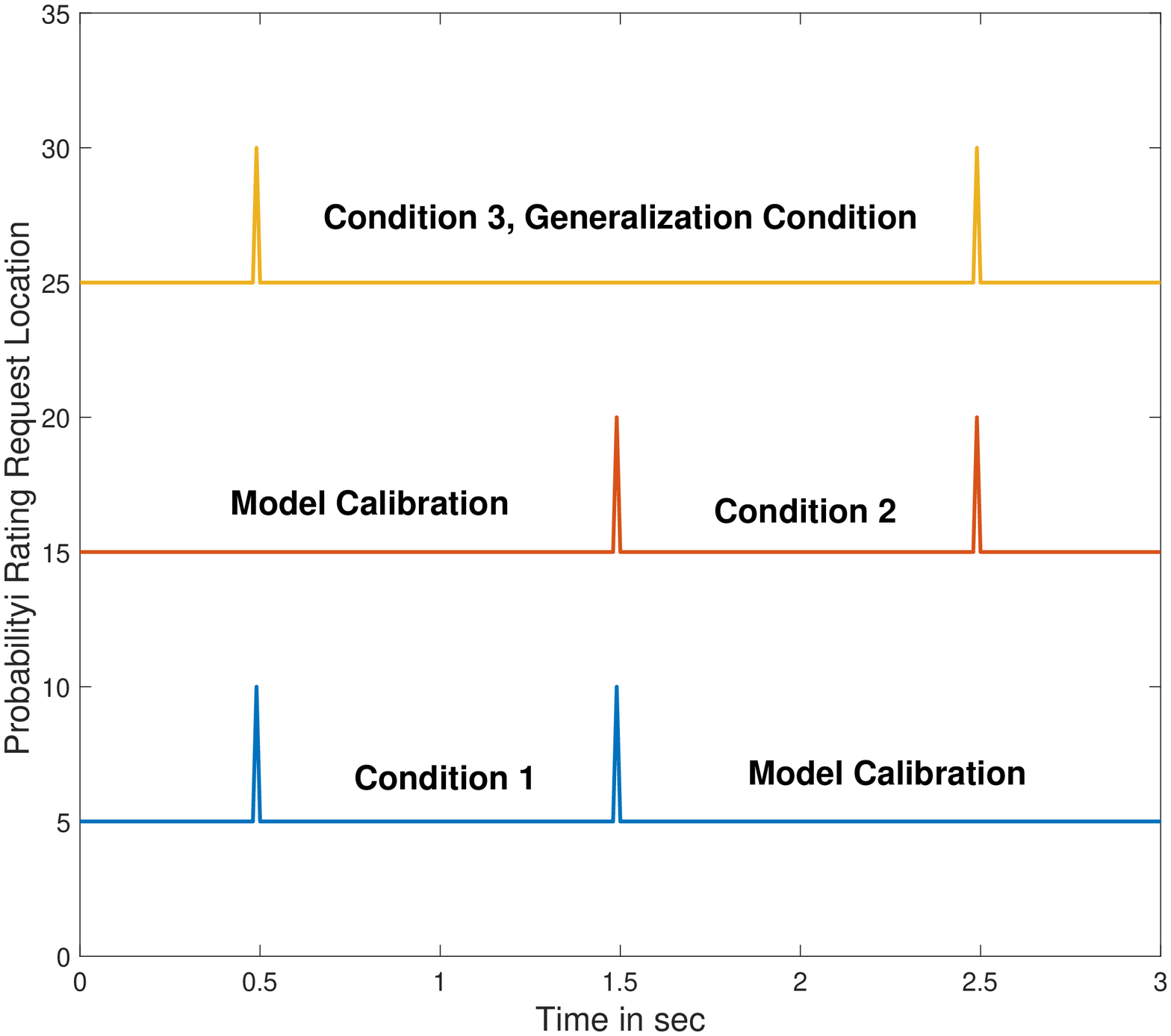}
\end{figure}

Previously, \cite{KvamPNAS2015} quantitatively compared the predictions
of these two models using a ``dot motion'' task for studying evidence
monitoring \cite{BallSekular1987}. This task has become popular
among neuroscientists for studying evolution of confidence \cite[see, e.g.,][]{ShadleKiani2009Conf}.
The dot motion task is a perceptual task that requires participants
to judge the left/right direction of dot motion in a display consisting
of moving dots within a circular aperture (see left panel of Figure
\ref{fig:Task1}). A small percentage of the dots move coherently
in one direction (left or right), and the rest move randomly. Difficulty
is manipulated between trials by changing the percentage of coherently
moving dots. The judge watches the moving dots for a period time at
which point the experimenter requests a probability rating for a direction
(see Figure \ref{fig:Task1}, left panel). In the study by \cite{KvamPNAS2015},
each of $9$ participants received over 2500 trials on the dot motion
task.

The experimental design used by \cite{KvamPNAS2015} included 4 coherence
levels (2\%, 4\%, 8\%, or 16\%) and two different kinds of judgment
conditions. In the choice-confidence condition, participants were
given $t_{1}=.5$ s to view the display, and then a tone was presented
that signaled the time to make a binary (left/right) decision. After
an additional $\Delta t=0.05,0.75,\ or\ 1.5s$ following the choice,
participants were prompted by a second tone to make a probability
rating on a $0$ (certain left) to $100$\% (certain right) rating
scale (see Figure \ref{fig:Task1}, middle panel). In a confidence-only
condition, participants didn't have to make any decision and instead
they simply made a pre-determined response when hearing the tone at
time $t_{1}$, and then later they made a probability rating at the
same total time points $t_{2}$ as the choice - confidence condition. 

According to a Markov model, the marginal distribution of confidence
at time $t_{2}$ (pooled across choices at time $t_{1}$ for the choice-confidence
condition) should be the same between the two conditions at time $t_{2}$
(see SI in \cite{KvamPNAS2015} for proof). According to the quantum
model, the decision at time $t_{1}$ produces a collapse, which introduces
interference during processing before the second judgment, and this
interference disturbs and changes the marginal distribution of confidence
for the choice-confidence condition at time $t_{2}$. The results
strongly favored the quantum model predictions: the interference effect
was significant at the group level, and at the individual level, 6
out of the 9 participants produced significant interference effects.
Furthermore, the Markov and quantum models were used to predict both
the binary choices and as well as the confidence ratings using the
same number of model parameters, and results of the model comparisons
strongly favored the quantum over the Markov model for 7 out of 9
participants.

The current paper presents a new type of comparison of the Markov
versus quantum models for the dynamics of belief change. The previous
study examined the sequential effects of a binary decision on a later
probability judgment; in contrast, the present study examines the
sequential effects of a probability judgment on a later judgment.
A binary decision may evoke a stronger commitment, whereas a probability
judgment does not force the decision maker to make any clear decision.
The question is whether the first probability judgment is sufficient
to produce an interference effect like that produced by committing
to a binary decision. 

A total of 11 participants (8 female, 3 male) were paid depending
on their performance for making judgments on approximately 1000 trials
across 3 daily sessions (see Methods for more details). Once again,
the participants monitored dot motion using 4 coherence levels (2\%,
4\%, 8\%, or 16\%) with half of the trials presenting left moving
dots and the remaining half of the trials presenting right moving
dots. In this new experiment, two probability ratings were made at
a pair $(t_{1},t_{2})$ of time points (see Figure \ref{fig:Task1},
right panel). The experiment included three main conditions: (1) requests
for probability ratings at times $t_{1}=.5s$ and $t_{2}=1.5s,$ (2)
requests for ratings at times $t_{1}=1.5s$ and $t_{2}=2.5s$, and
(3) requests for ratings at times $t_{1}=.5s$ and $t_{2}=2.5s$.
This design provided two new tests of Markov and quantum models. 

First of all, we tested for interference effects by comparing the
marginal distribution of probability ratings at time $t_{2}=1.5s$
for condition 1 (pooled across ratings made at time $t_{1}=.5s$ )
with the distribution of ratings at time $t_{1}=1.5s$ from condition
2. Once again, the Markov model predicts no difference between conditions
at the matching time points, whereas the quantum model predicts an
interference effect of the first rating on the second.

Secondly, and more important, this design provided a new generalization
test \cite{BusemeyerWang2000} for quantitatively comparing the predictions
computed from the competing models. The generalization test provides
a different and potentially more robust method than the previously
used Bayes factor for comparing the two models because it is based
on \emph{a priori} predictions made to new conditions. The parameters
from both models were estimated from the probability ratings distributions
obtained from the first two conditions for each individual; then these
same parameters were used to predict probability rating distribution
for each person on the third condition (see Figure \ref{fig:Task1}).
Both models used two parameters to predict the probability rating
distributions (see Methods for details): one that we call the ``drift''
rate that affects the direction and strength of change in the distribution
of beliefs, and another that we call the ``diffusion'' rate that
affects the speed of change and dispersion of the distributions. Using
maximum likelihood (see Methods for details), we estimated these two
parameters from the joint distribution (pair of ratings at $.5s$
and $1.5s$) obtained from condition 1, and the joint distribution
(pair of ratings at $1.5s$ and $2.5s$) from condition 2, separately
for each coherence level and each participant. Then we used these
same two parameters to predict the joint distribution (pair of ratings
$.5s$ and $2.5s$) obtained from condition 3 for each coherence level
and participant.

\section*{Results}

The probability ratings were made by moving a cursor (via joystick)
across the edge of a semi-circular scale ranging from 0 (certain moving
left) to 100 (certain moving right). Ratings for right-moving dots
were used directly; but ratings for left-moving dots were rescored
as (100 - rating). In this way, a rating of zero represented certainty
that dots were moving in the incorrect direction, a rating of 50 represented
uncertainty about the direction, and a rating of 100 represented certainty
that the dots were moving in the correct direction. 

The mean (standard deviation) probability rating at time $t_{1}$,
pooled across conditions, equaled ($53.84(4.07)$, $59.51(6.33)$, $66.60(11.08)$, $79.72(16.81)$)
for coherence levels 2\%,4\%,8\%,16\%, respectively. The prediction
of interference derived from the quantum model relies on the assumption
that there is second stage processing of information; without second
stage processing, the quantum model predicts no interference (cf.,
\cite{KvamPNAS2015}). To check this assumption, we tested the
effect of the second stimulus interval on the change in probability
ratings from time $t_{1}$ to $t_{2}$ by computing the mean change
for each person and coherence level (averaged over conditions). The
mean (standard deviation) change across participants equaled $1.28(1.87),1.03(2.94),2.75(2.42),2.01(1.86)$
for coherence levels 2,4,8, and 16 respectively. According to a Hotelling
T test, this vector of change is significantly different from zero
($F(3,8)=4.5765,p=.039$), indicating that participants' judgments
moved toward response values in favor of the correct dot motion direction
over the second time interval, on average.

\begin{figure}
\caption{\label{fig:RatingDist}Relative frequency distribution of ratings
at 2\% coherence level for conditions 1 at $t_{2}=1.5$ and condition
2 at $t_{1}=1.5$, and interference effect between these conditions.
The horizontal axis represents the probability ratings for the correct
direction, and the vertical axis represents the proportion assigned
to each rating value. Top panel shows results for condition 1, middle
panel shows condition 2, and bottom panel shows the difference (top
minus middle).}
\begin{centering}
\includegraphics[width=.75\textwidth]{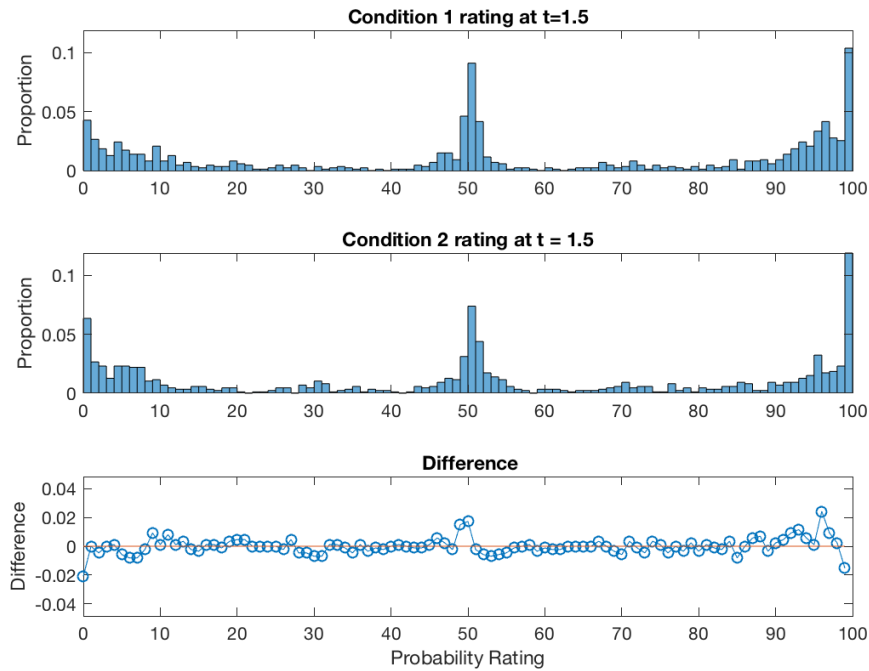}
\end{centering}

\end{figure}

Figure \ref{fig:RatingDist} shows the relative frequency distribution
of ratings for the lowest (2\%) coherence level for conditions 1 at
$t_{2}=1.5s$ and condition 2 at $t_{1}=1.5s$, and the difference
between the two. First note that the ratings tended to cluster into
three groups near the end points and middle point of the probability
scale. This clustering also occurred with all of the other coherence
levels and across participants. Based on the finding that the ratings
tended to cluster into three groups, we categorized the data into
three levels (L=low = ratings from $0$ to 33, M=medium = ratings
from 34 to 66, and H=high = ratings from 67 to 100). This also had
the benefit of increasing the frequencies within the cells, which
was required for the chi - square statistical tests reported next. 

\subsection*{Statistical Tests of Treatment Effects}

First we statistically tested for an interference effect between conditions
1 and 2 using the categorized ratings. For this test, we compared
the marginal distribution across the three categories for condition
1 at time $t_{2}=1.5s$ with the marginal distribution across the
three categories for condition 2 at time $t_{1}=1.5s$. The chi square
difference between the marginals for the two conditions was first
computed separately for each participant and coherence level, and
then summed across participants for each coherence level to produce
a total chi square test at each coherence level. The results produced
significant differences only for the low coherence levels (the $G^{2}$
(chi square statistics) are 38.0, 27.6, 25.4, 28.1, for coherence
levels 2\%, 4\%, 8\%, and 16\% respectively, and the critical value
for $\alpha=.05$ and $df=11\cdot(3-1)=22$ equals $33.9$). Only
$3$ out of the $11$ participants produced significant effects at
the low (2\%, 4\%) coherence levels. 

Second, we statistically tested the difference between the joint probability
distributions for conditions 1 versus 3 and again for conditions 2
versus 3. For the test between condition 1 versus 3, we compared the
$3\times3$ joint distribution produced by category ratings at time
$t_{1}$ and $t_{2}$ for condition 1 with the $3\times3$ joint distribution
produced by category ratings at time $t_{1}$ and $t_{3}$ for condition
3; likewise for the test comparing conditions 2 and 3. The chi square
difference between two $3\times3$ joint distributions was first computed
for each person separately, and then summed across participants. The
results produced significant differences for both the condition 1
versus 3 comparison ($G^{2}$ = 116.9, $df=(9-1)\cdot11=88,$ $p=.0215$)
and for the condition 2 versus 3 comparison ($G^{2}$ = 192.4, $df=(9-1)\cdot11=88,$
$p<.0001$). Five of the 11 participant produced significant differences. 

In summary, the results suggest that interference effects do occur
with sequences of judgments, but they are small and occur for only
a subset of the participants and coherence conditions. The results
also show differences between the calibration conditions (1,2) and
the generalization condition 3.

\subsection*{Model comparisons}

A more powerful test of the Markov versus quantum models was performed
for each participant using the generalization criterion method \cite{BusemeyerWang2000}.
Both models have two parameters (a drift and a diffusion parameter,
see Methods). These two parameters were estimated, based on maximum
likelihood, separately for each participant and each coherence level
using the data from the pair of $3\times3$ joint distributions produced
by responses in conditions 1,2. Then these same parameters were used
to compute the predictions for each participant and coherence level
for condition 3. The discrepancy between data and model predictions
was measured using a $G^{2}=-2\cdot log\:likhood$ chi-square statistic
(see Methods), and we computed the difference between models defined
as $G_{diff}^{2}=G_{Markov}^{2}-G_{quantum}^{2}$. The $G_{diff}^{2}$
statistics, summed across participants, were $350,306,260,40$ for
coherence levels 2\%,4\%,8\%,16\% respectively, favoring the quantum
model over the Markov model. Eight of the $11$ participants produced
$G_{diff}^{2}$ favoring the quantum model for coherence levels 2\%,
4\%, and 8\%, but only 5 participants produced results favoring the
quantum model for coherence level 16\%. The results clearly favor
the quantum model, but less so for high coherence. 

Tables \ref{tab:GenTest1}-\ref{tab:GenTest4} show the predicted
and observed frequencies of responses (3x3 tables), averaged across
participants, for each coherence level. The observed data reveals
large frequencies at both Low and High ratings under low coherence
conditions. The wave nature of the quantum model provides a way to
spread the judgments across both Low and High levels. However, the
sample path nature of the Markov model makes it difficult to simultaneously
distribute frequencies to both Low and High ratings. To address this
problem with the Markov model, a revised Markov model, shown as Markov-V
in the tables, is discussed next. 

\begin{table}
\centering
\caption{\label{tab:GenTest1} Observed and predicted distributions, averaged
across participants, for coherence level 1. }

\begin{tabular}{cccc|ccc|ccc|ccc}
 & \multicolumn{3}{c}{Obs} & \multicolumn{3}{c}{Markov} & \multicolumn{3}{c}{Quantum} & \multicolumn{3}{c}{Markov-V}\tabularnewline
 & L2 & M2 & \multicolumn{1}{c}{H2} & L2 & M2 & \multicolumn{1}{c}{H2} & L2 & M2 & \multicolumn{1}{c}{H2} & L2 & M2 & H2\tabularnewline
L1 & .20 & .02 & .07 & .09 & .07 & .04 & .13 & .05 & .07 & .26 & .01 & .01\tabularnewline
M1 & .04 & .23 & .05 & .11 & .27 & .16 & .04 & .26 & .08 & .08 & .18 & .09\tabularnewline
H1 & .05 & .02 & .32 & .03 & .08 & .16 & .06 & .04 & .25 & .01 & .01 & .35\tabularnewline
\end{tabular}

\caption{\label{tab:GenTest2} Observed and predicted distributions, averaged
across participants, for coherence level 2. }

\begin{tabular}{cccc|ccc|ccc|ccc}
 & \multicolumn{3}{c}{Obs} & \multicolumn{3}{c}{Markov} & \multicolumn{3}{c}{Quantum} & \multicolumn{3}{c}{Markov-V}\tabularnewline
 & L2 & M2 & \multicolumn{1}{c}{H2} & L2 & M2 & \multicolumn{1}{c}{H2} & L2 & M2 & \multicolumn{1}{c}{H2} & L2 & M2 & H2\tabularnewline
L1 & .17 & .02 & .07 & .06 & .06 & .04 & .14 & .05 & .04 & .21 & .01 & .01\tabularnewline
M1 & .02 & .22 & .05 & .08 & .25 & .18 & .04 & .25 & .09 & .06 & .17 & .09\tabularnewline
H1 & .05 & .02 & .38 & .03 & .08 & .22 & .03 & .04 & .33 & .01 & .02 & .42\tabularnewline
\end{tabular}

\caption{\label{tab:GenTest3} Observed and predicted distributions, averaged
across participants, for coherence level 3. }

\begin{tabular}{cccc|ccc|ccc|ccc}
 & \multicolumn{3}{c}{Obs} & \multicolumn{3}{c}{Markov} & \multicolumn{3}{c}{Quantum} & \multicolumn{3}{c}{Markov-V}\tabularnewline
 & L2 & M2 & \multicolumn{1}{c}{H2} & L2 & M2 & \multicolumn{1}{c}{H2} & L2 & M2 & \multicolumn{1}{c}{H2} & L2 & M2 & H2\tabularnewline
L1 & .12 & .02 & .04 & .04 & .04 & .04 & .12 & .04 & .03 & .16 & .01 & .01\tabularnewline
M1 & .02 & .18 & .07 & .05 & .21 & .18 & .04 & .24 & .09 & .05 & .15 & .07\tabularnewline
H1 & .03 & .02 & .50 & .02 & .08 & .34 & .02 & .05 & .38 & .01 & .01 & .53\tabularnewline
\end{tabular}

\caption{\label{tab:GenTest4} Observed and predicted distributions, averaged
across participants, for coherence level 4. }

\begin{tabular}{cccc|ccc|ccc|ccc}
 & \multicolumn{3}{c}{Obs} & \multicolumn{3}{c}{Markov} & \multicolumn{3}{c}{Quantum} & \multicolumn{3}{c}{Markov-V}\tabularnewline
 & L2 & M2 & \multicolumn{1}{c}{H2} & L2 & M2 & \multicolumn{1}{c}{H2} & L2 & M2 & \multicolumn{1}{c}{H2} & L2 & M2 & H2\tabularnewline
L1 & .05 & .01 & .03 & .02 & .01 & .02 & .04 & .01 & .01 & .06 & .01 & .00\tabularnewline
M1 & .01 & .14 & .05 & .03 & .13 & .11 & .01 & .17 & .04 & .03 & .12 & .05\tabularnewline
H1 & .02 & .01 & .68 & .01 & .05 & .63 & .01 & .03 & .67 & .00 & .01 & .71\tabularnewline
\end{tabular}

Note: For example, L1 stands for Low rating after first interval and
M2 stands for High rating after second interval. Cells in the upper
right off diagonal represent transitions from lower to higher probability
ratings during the second interval.
\end{table}

One possible reason for the lower performance of the Markov is that
it does not include any variability in the drift rate across trials.
It has been argued that drift rate variability is required to produce
accurate fits for the Markov model (see, e.g., \cite{RatcliffTiCS2016}).
In fact, the actual proportion of coherent dots varied across trials
within a single coherence level, because the coherent dots were randomly
sampled from a Binomial distribution with $N=70$ dots. This produced
a stimulus that varied from trial to trial, meaning that the drift
rate of the evidence accumulation models should also shift from trial
to trial. To allow for this variability, we recomputed the predictions
of the Markov model by averaging the predictions across the values
$0$ to $1$ in steps of 1/70 assuming a Binomial distribution with
$N=70$, where the mean of the Binomial distribution was then a parameter
that was fit for each participant and each coherence level (along
with the diffusion rate parameter, see Methods). This revised Markov
model produced $G_{diff}^{2}$ statistics, summed across participants,
equal to $198,165,-26,-130$ for coherence levels 2\%,4\%,8\%,16\%
respectively (once again, positive values indicate evidence for the
quantum model and negative values indicate evidence for the Markov
model). The quantum model is favored for the lower coherence levels,
but the Markov-V model is favored for the higher coherence levels.
Six of the $11$ participants produced $G_{diff}^{2}$ favoring the
quantum model for coherence levels 2\% and 4\%, four of the 11 participants
favored the quantum model for coherence level 8\%, but only 1 produced
results favoring the quantum model for coherence level 16\%. Now the
results only favor the quantum model for low coherence, and they favor
the Markov model for high coherence. 

The prediction of the Markov model with drift rate variability are
also shown on the right side of Tables \ref{tab:GenTest1}-\ref{tab:GenTest4}.
The predictions of the Markov-V model are much improved over the original
Markov model, and the accuracy of the Markov-V model is now comparable
to the quantum model.



\section*{Discussion}
This article empirically evaluated two different types of dynamic
models for belief change during evidence monitoring.  According to a Markov process, the
decision maker's belief state acts like a particle that changes from
one location to another producing a sample path across time. 
In contrast, according to the quantum model, the decision maker's
belief state is like a wave spread across the evidence scale that
flows across time. 
These two competing models can be compared using both qualitative
tests of properties of each models as well as quantitative comparisons
of predictive accuracy.

The Markov and quantum processes make different predictions regarding
interference effects that can occur when a sequence of responses are
requested from the decision maker. As mentioned earlier, \cite{KvamPNAS2015}
examined interference effects under
a ``decide and then judge'' condition. That earlier experiment
produce significant interference effects such that confidence was
less extreme following a binary decision, and the size of the interference
was directly related to the size of the effect of second stage processing,
as predicted by the quantum model. The present examined interference effects under
a ``judge and then judge'' condition. This new
experiment indicated that an interference effect did occur at the
low levels of confidence, but the effect was small and only occurred
with 3 out 11 participants. One way to interpret this difference in
empirical results is that using a binary decision for the first measurement
may be more effective for ``collapsing'' the wave function than
using a probabilistic judgment for the first measurement, resulting
in greater interference between choice and confidence responses than
for sequential confidence responses.

The present experiment is unique in the way the two models were quantitatively
compared by using a more powerful generalization criterion method,
which allowed us to examine not just how well the models fit the data,
but how well they could predict new data. Using this method, the parameters
of the models were estimated from conditions 1,2 and these same parameters
were used to predict a new condition 3. The results of the present
experiment indicated that the quantum model produced more accurate
predictions for low levels of confidence, but the Markov-V model (with
drift rate variability) produced more accurate predictions for high
levels of confidence. Together these results suggest that neither
process alone, quantum or Markov, is sufficient to account all conditions
and all participants.

Rather than treating Markov and quantum models as mutually exclusive,
an alternative idea is that a more general hybrid approach is needed,
one that integrates both quantum and Markov processes. The quantum
model used in the present work is technically viewed as a ``closed
system'' quantum process with no external environmental forces. However,
it is possible to construct an \textquotedbl open system\textquotedbl{}
quantum process where a person's state partially decoheres as a result
of interaction with a noisy mental environment. A coherent way to
accomplish a combined quantum-Markov process can be formed by using
a more general open system quantum process that includes a process
representing these external environmental forces \cite{martinez2016quantum,Accardi2009,khrennikova2016quantum}.
Open system quantum models start out in a coherent quantum regime,
and later decohere into a classical Markov regime \cite[see, e.g.,][]{Accardi2009}.
In fact, \cite{FussNavaro2013} compared Markov and quantum models
with respect to their predictions for both choice and decision time:
When a closed system quantum model was compared to the Markov model,
there was a slight advantage for the Markov model; however, when an
open system quantum model was used, the quantum model produced a small
advantage. For our application, we would need to speculate that the
speed of decoherence depends on coherence level, but the development
of a specific open system quantum model for belief change is left
for future research.

\section*{Methods}


\subsection*{Participants}

A total of 11 Michigan State University (8 female, 3 male) students
were recruited for the study -- 1 additional participant began the
study but was dropped for failing to complete all sessions of the
study. Each of the 11 remaining participants completed 3 sessions
of the study, and were paid \$10 per session plus an additional bonus
based on the accuracy of their confidence ratings -- up to \$5 based
on how close they were to the ``100\% confident in the correct direction''
responses on each trial. Each participant competed approximately 1000
trials of the task across all sessions.

All methods were carried out in accordance with relevant guidelines and regulations, 
and all experimental protocols were approved by the Michigan State Human Subjects Review Board, and informed consent was obtained from all subjects.

\subsection*{Task}

In the task, participants viewed a random dot motion stimulus where
a set of dots were presented on screen. Most of these dots moved in
random directions, but a subset of these dots were moving coherently
to either the left or the right side of the screen. The dots were
white dots on a black background which composed a circular aperture
of approximately 10 visual degrees in diameter. The display was refreshed
at 60 Hz and dots were grouped into 3 dot groups that were presented
in sequence (group 1, 2, 3, 1, 2, 3, \dots ) and displaced by a quarter
of a degree every time they appeared on screen, for apparent motion
at 5 degrees per second. When prompted, participants indicated their
confidence that the dots were moving left or right on a scale from
0 (certain that they are moving left) to 100 (certain that they are
moving right). They entered their responses by using a joystick to
move the cursor across the edge of a semicircular confidence scale
like the one shown in Figure \ref{fig:Task1}.

To begin each trial, participants pressed the trigger button on a
joystick in front of them while the cursor -- presented as a crosshair
-- was in the middle of the screen. The random dot stimulus then
appeared on the screen, with 2\%, 4\%, 8\%, or 16\% of the dots moving
coherently toward one (left vs. right) direction. After 500 ms or
1500 ms, participants were prompted for their first probability judgment
response with a 400 Hz auditory beep. They responded by moving the
cursor across the semicircular confidence scale at the desired probability
response. Since participants were using a joystick, the cursor naturally
returned to the center of the screen after this initial response.
Once the first response had been made, the stimulus remained for an
additional 1000 or 2000 ms before a second auditory beep prompting
the second probability response. Participants made their second response
in the same way as the first. This resulted in a possible on-screen
stimulus time of 1500 or 2500 ms plus the time it took to respond
(it was never the case that $t_{1}$ = 1500 and $t_{2}$ = 2000 ms).

After each trial, participants received feedback on what the correct
dot motion direction was and how many points they received for their
confidence responses on that trial. We recorded the amount of time
it took participants to respond after each auditory beep, the confidence
responses they entered, the number of points received for the trial,
and the stimulus information (coherence, direction, beep times). Everything
was presented and recorded in Matlab using Psychtoolbox and a joystick
mouse emulator \cite{kleiner2007s}. 

\subsection*{Procedure}

Participants volunteered for the experiment by signing up through
the laboratory on-line experiment recruitment system, which included
mainly Michigan State students and the East Lansing community. Upon
entering the lab, they completed informed consent and were briefed
on the intent and procedures of the study. The first experimental
session included extensive training on using the scale and joystick,
including approximately 60 practice trials on making accurate responses
to specific numbers, single responses to the stimulus, making two
accurate responses to numbers in a row, and making two responses to
the stimulus (as in the full trials).

Subsequent experimental sessions started with 30-40 ``warm-up\textquoteright \textquoteright{}
trials that were not recorded. After training or warm-up, participants
completed 22 (first session) or 28 (subsequent sessions) blocks of
12 trials, evenly split between confidence timings and stimulus coherence
levels. After every block of trials, they completed 3 test trials
where they were asked to hit a particular number on the confidence
scale rather than respond based on the stimulus. This was included
to get a handle on how accurate and precise the participants could
be when using the joystick and understand how much motor error was
likely factoring into their responses.

At the conclusion of the experiment, participants were debriefed on
its intent and paid \$10 plus up to \$5 according to their performance.
Performance was assessed using a strictly proper scoring rule \cite{MerkleSteyvers2103}
so that the optimal response was to give a confidence response that
reflected their expected accuracy. Participants received updates on
the number of points they received at the end of each block of the
experiment, including at the end of the study.

\subsection*{Mathematical Models}

There are different types of Markov processes that have been used
for evidence accumulation. One type is a discrete state and time Markov
chain \cite{busemeyer1982choice} , another type is a discrete state
and continuous time random walk process \cite{PikeMarkov}, another
type is a continuous state and discrete time random walk \cite{Laming1968,link1975relative},
and third type is a continuous state and continuous time diffusion
process \cite{Ratcliff1978}. However, the discrete state converges
to make the same predictions as the diffusion process when there are
a large number of states and the step size approaches zero \cite[see][]{DiederichBusemMatrixMeth2003}.

Like the Markov models, there are different types of quantum processes.
One type is a discrete state continuous time version \cite{KvamPNAS2015,martinez2016quantum}
and another type is a continuous state and time version \cite{Busemeyer:Wang:Townsend:2006}. 

To facilitate the model comparison, we tried to make parallel assumptions
for the two models. The use of a discrete state and continuous time
version for both the Markov and quantum models serves this purpose
very well. Additionally, a large number of states were used to closely
approximate the predictions of continuous state and time processes.

For both models, we used an approximately continuous set of mental
belief states. The set consisted of $N=99$ states $j\in\{1,...,99\}$,
where $1$ corresponds to a belief that the dots are certainly not
moving to the right (i.e., a belief that they are certainly moving
to the left), 50 corresponds to completely uncertain belief state,
and 99 corresponds to a belief that the dots are certainly moving
to the right. We used 1-99 states instead of $0-100$ states because
we categorized the states into three categories and $99$ can be equally
divided into three sets. For a Markov model, the use of $N=99$ belief
states produces a very closely approximation to a diffusion process. 

For the Markov model, we define $\phi_{j}(t)$ as the probability
that an individual is located at a belief state $j$ at time $t$
for a single trial, which is a positive real number between $0$ and
$1,$ and $\sum\phi_{j}(t)=1$. These 99 state probabilities form
a $N\times1$ column matrix denoted as $\phi(t)$. For the quantum
model, we define $\psi_{j}$ as the amplitude that an individual assigns
to the belief state located a evidence level $j$ on a single trial
(the probability of selecting that belief state equals $\left|\psi_{j}\right|^{2}$).
The amplitudes are complex numbers with modulus less than or equal
to one, and $\sum\left|\psi\right|^{2}=1$. Both models assumed a
narrow, approximately normally distributed (mean zero, standard deviation
= 5 steps in the 99 states), initial probability distribution at the
start ($t=0)$ of each trial of the task. 

The probability distribution for the Markov process evolves from $\tau$
to time $\tau+t$ according to the transition law $\phi(t+\tau)=T(t)\cdot\phi(\tau)$,
where $T(t)$ is a transition matrix defined by the matrix exponential
function $T(t)=exp(t\cdot K)$. Transition matrix element $T_{ij}$
is the probability to transit from the state in column $j$ to the
state in row $i.$ The intensity matrix $K$ is a $N\times N$ matrix
defined by matrix elements $K_{ij}=\alpha>0$ for $i=j-1$, $K_{ij}=\beta>0$
for $i=j+1$, $K_{ii}=-\alpha-\beta$, and zero otherwise. The amplitude
distribution for the quantum process evolves from $\tau$ to time
$\tau+t$ according to the unitary law $\psi(t+\tau)=U(t)\cdot\psi(\tau)$,
where $U(t)$ is a unitary matrix defined by the matrix exponential
function $U(t)=exp(-i\cdot t\cdot H)$. Unitary matrix element $U_{ij}$
is the amplitude to transit from the state in column $j$ to the state
in row $i.$ The Hamiltonian matrix $H$ is a $N\times N$ Hermitian
matrix defined by matrix elements $H_{ij}=\sigma$ for $i=j+1$, $H_{ij}=\sigma^{*}$
for $i=j-1$, $H_{ii}=\mu\cdot\frac{i}{N}$, and zero otherwise.

For both models, we mapped the 99 belief states to 3 categories using
the following three orthogonal projection matrices $M_{L},$ $M_{M},$
and $M_{H}$. Define $\mathbf{1}$ as a vector of 33 ones, and define
$\mathbf{0}$ as a vector of 33 zeros. Then $M_{L}=diag[\mathbf{1},\mathbf{0},\mathbf{0}]$,
$M_{M}=diag[\mathbf{0},\mathbf{1},\mathbf{0}]$ $M_{H}=diag[\mathbf{0},\mathbf{0},\mathbf{1}]$.
Finally, define $\left\Vert X\right\Vert ^{1}$ as the sum of all
the elements in the vector $X,$ and $\left\Vert X\right\Vert ^{2}$
as the sum of the squared magnitude of the elements in the vector
$X.$ For the Markov model, the joint probability of choosing category
$k$ at time $t_{1}$ and then choosing category $l$ at time $t_{2}$
equals 
\begin{equation}
p(R(t_{1})=k,R(t_{2})=l]=\left\Vert M_{l}\cdot T(t_{2}-t_{1})\cdot M_{k}\cdot T(t_{1})\cdot\phi(0)\right\Vert ^{1}.
\end{equation}
For the quantum model, the joint probability of choosing category
$k$ at time $t_{1}$ and then choosing category $l$ at time $t_{2}$
equals 
\begin{equation}
p(R(t_{1})=k,R(t_{2})=l]=\left\Vert M_{l}\cdot U(t_{2}-t_{1})\cdot M_{k}\cdot U(t_{1})\cdot\psi(0)\right\Vert ^{2}.
\end{equation}

The Markov model required fitting two parameters: a ``drift'' rate
parameter $\mu=\frac{\alpha}{\alpha+\beta}$ and a diffusion rate
parameter $\gamma=(\alpha+\beta)$. The quantum model required fitting
two parameters: a ``drift'' rate parameter $\mu$ , and a ``diffusion''
parameter $\sigma$. The parameter $\mu$ must be real, but $\sigma$
can be complex. However, to reduce the number of parameters, we forced
$\sigma$ to be real. The model fitting procedure for both the Markov
and the quantum models entailed estimating the two parameters from
conditions 1 and 2 separately for each participant and each coherence
level using maximum likelihood. 

The Markov-V model used a binomial distribution of ``drift'' rate
parameter $\mu\sim Bin(N=70,\upsilon$) and $\upsilon$ was estimated
using maximum likelihood. The predictions for the Markov-V model were
then obtained from the expectation 
\begin{equation}
p(R(t_{1})=k,R(t_{2})=l]=\sum_{n=0}^{n=70}p(\mu=\frac{n}{70})\cdot p\left[R(t_{1})=k,R(t_{2})=l|\mu\right]
\end{equation}
 
\section*{Data availability} The datasets and computer programs used in the current study are available at http://mypage.iu.edu/~jbusemey/quantum/DynModel/DynModel.htm.

\bibliography{Busemeyer}



\section*{Acknowledgements}
This research was supported by AFOSR Grant FA9550 15 1 0343 to JRB, an NSF Grant SBE 0955410 to TJP, and an NSF graduate research fellowship DGE 1424871 to PDK

\section*{Author Contributions} 
T.P. and P.K. conceived and conducted the experiment. J.B. analyzed results. All three helped write manuscript.

\section*{Additional information}

\textbf{Competing interests} 
The author(s) declare no competing interests.. 





\end{document}